\documentclass[letterpaper]{article} 
\usepackage{aaai2026}  
\usepackage{times}  
\usepackage{helvet}  
\usepackage{courier}  
\usepackage[hyphens]{url}  
\usepackage{graphicx} 
\usepackage{natbib}  
\usepackage{caption} 
\usepackage{algorithm}
\usepackage{algorithmic}

\usepackage{newfloat}
\usepackage{listings}
\usepackage{amsmath}
\usepackage{booktabs}
\usepackage{multirow}
\usepackage{amssymb}
\usepackage{array}
\DeclareCaptionStyle{ruled}{labelfont=normalfont,labelsep=colon,strut=off} 
\floatstyle{ruled}
\newfloat{listing}{tb}{lst}{}
\floatname{listing}{Listing}
\title{CaTFormer: Causal Temporal Transformer with Dynamic Contextual Fusion for Driving Intention Prediction}
\author{
    Sirui Wang\equalcontrib,
    Zhou Guan\equalcontrib,
    Bingxi Zhao,
    Tongjia Gu,
    Jie Liu\thanks{Corresponding Author.}
}
\affiliations{
    Beijing Jiaotong University\\
    \{siruiwang, zhouguan, bingxizhao, gut2gu, jieliu\}@bjtu.edu.cn
}

\begin{document}
\maketitle

\begin{abstract}
Accurate prediction of driving intention is key to enhancing the safety and interactive efficiency of human-machine co-driving systems. It serves as a cornerstone for achieving high-level autonomous driving. However, current approaches remain inadequate for accurately modeling the complex spatiotemporal interdependencies and the unpredictable variability of human driving behavior. To address these challenges, we propose CaTFormer, a causal Temporal Transformer that explicitly models causal interactions between driver behavior and environmental context for robust intention prediction. Specifically, CaTFormer introduces a novel Reciprocal Delayed Fusion (RDF) mechanism for precise temporal alignment of interior and exterior feature streams, a Counterfactual Residual Encoding (CRE) module that systematically eliminates spurious correlations to reveal authentic causal dependencies, and an innovative Feature Synthesis Network (FSN) that adaptively synthesizes these purified representations into coherent temporal representations. Experimental results demonstrate that CaTFormer attains state-of-the-art performance on the Brain4Cars dataset. It effectively captures complex causal temporal dependencies and enhances both the accuracy and transparency of driving intention prediction.
\end{abstract}

\begin{links}
    \link{Code}{https://github.com/srwang0506/CaTFormer}
\end{links}

\section{Introduction}
Driver intention prediction is crucial for autonomous driving systems, as it effectively mitigates risks and enhances driving safety. By forecasting potential outcomes several seconds in advance, the system can proactively alert the driver or initiate evasive maneuvers, significantly improving its safety capabilities.

Initially, intention prediction primarily relied on the extraction and fusion of visual features for basic predictions \cite{huang2022survey}. However, advancements in sensor technology have enabled the incorporation of multi-modal information, such as GPS coordinates, vehicle speed, map data, and driver head pose \cite{hu2021data,mo2023map,li2022pomdp,wu2023graph}. By leveraging complex models for feature extraction, fusion, and prediction, the performance of driver intention prediction has significantly improved \cite{sui2021joint,TIFN2023,liu2023real,gao2023dual}.

\begin{figure}[!t]
    \centering
    \includegraphics[width=1\linewidth]{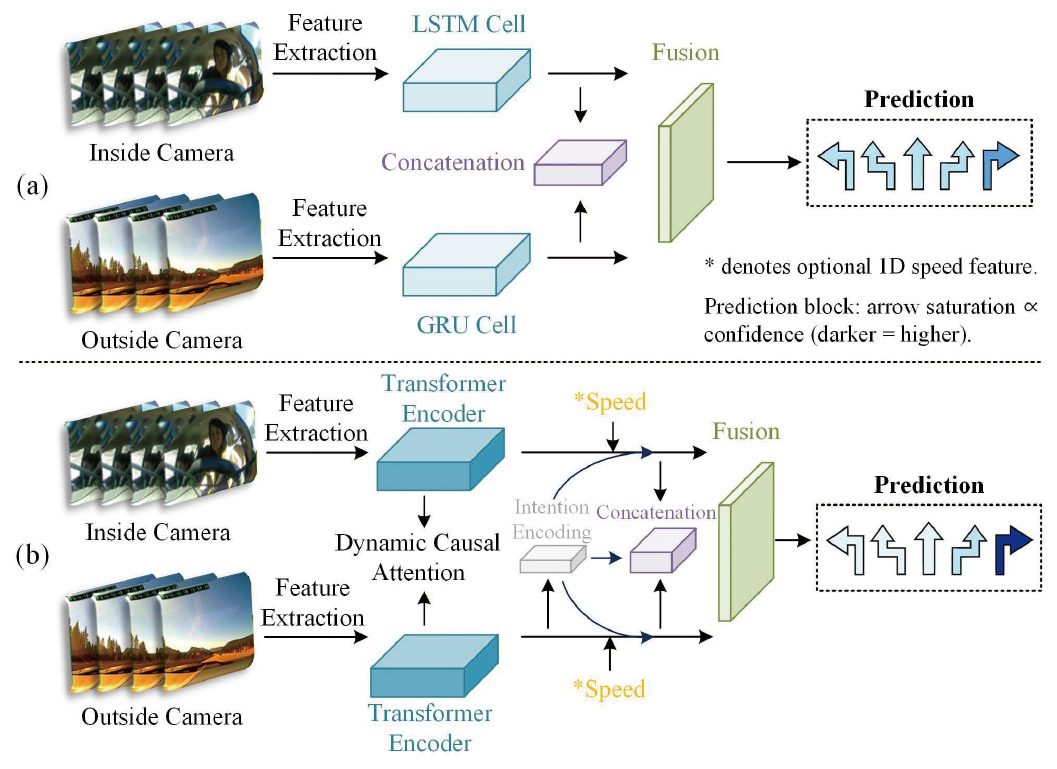}
    \caption{Comparison between previous driving intention prediction methods and ours. (a) is an LSTM-GRU framework processing interior and exterior streams independently before concatenation. (b) is our CaTFormer, a Transformer-based model enabling dynamic causal fusion of dual streams with integrated intention priors. Through joint modeling of global-local dependencies and cross-stream interactions, our approach outperforms existing methods.
}
    \label{fig:intro}
\end{figure}

Despite advancements in driver intention prediction, the rich multi-modal data remains underutilized. Most existing studies simply concatenate or linearly aggregate this information, as shown in Fig. \ref{fig:intro} (a). However, given that the driver controls the vehicle, changes in their state directly influence the vehicle's driving status, indicating a strong dependency. Therefore, we propose to explicitly model the causal relationship between the driver and the environment, highlighting this causality's decisive impact on the prediction task, as illustrated in Fig. \ref{fig:intro} (b).

Specifically, we adopt a Transformer-based architecture and introduce three sequential components to enhance the causal modeling, encode the driver's intention, and fuse the multi-dimensional features. First, we introduce a Reciprocal Delayed Fusion (RDF) module that cross-fuses interior and exterior features through a shifting mechanism, explicitly establishing a temporal dependency between the two feature streams. As the fused features may contain considerable causal noise, we further devise a Counterfactual Residual Encoding (CRE) module to filter out such noise to obtain a more explicit causal representation. Finally, we utilize a Feature Synthesis Network (FSN) that employs a gating mechanism to integrate the interior, exterior, and interaction representations, enabling the modeling of both local and global causal structures.
Our main contributions are as follows:

\begin{itemize}
    \item We propose CaTFormer, an efficient Transformer-based framework for driving intention prediction that embeds causal spatio‑temporal reasoning with adaptive multi‑view fusion in a unified end‑to‑end architecture.
    \item Through dual-stream reciprocal delayed fusion, CaTFormer explicitly captures dependencies across interior and exterior streams, isolates genuine causal effects via counterfactual attention subtraction, and adaptively integrates complementary visual cues, effectively enhancing robustness under complex driving conditions.
    \item Extensive evaluation on the Brain4Cars dataset demonstrates that CaTFormer demonstrates superior performance in driving intention prediction in both highway and urban scenarios.
\end{itemize}

\section{Related Work}
In the early stages of driving intention prediction research, studies primarily focused on learning spatiotemporal representations directly from raw video, often employing 3D CNN-LSTM architectures for maneuver prediction \cite{Patrick2019}. While some later work attempted to enrich context by fusing interior and exterior streams via convolutional LSTMs for a more comprehensive decision-making basis \cite{Yao2020}, both approaches struggled with limited capacity for modeling long-range, non-consecutive dependencies.

Inspired by human cognitive processes, TIFN \cite{TIFN2023} introduced a state update unit (STU) to integrate environmental context into driver state modeling and extract semantic segmentation features as attention cues. Similarly, another study framed intention prediction as a sequence-labeling task, combining bidirectional LSTMs with a conditional random field to capture the contextual dependencies of driving behaviors \cite{Zhou2021}. Although these methods incorporate multi-source information, they typically learn inter-modal feature correlations implicitly, lacking explicit disentanglement and reasoning of their interactions. Moreover, to enhance model generalization, existing studies have developed personalized prediction models using techniques like domain-adversarial RNN \cite{Michele2019}, inverse reinforcement learning \cite{IDIPN2025}, and a federated learning framework \cite{FedPRM2024}.

The Transformer architecture excels at capturing long-range dependencies and global interaction information in temporal data through its unique attention mechanism \cite{liu2024laformer}. Building on this, CemFormer \cite{ma2023cemformer} integrates data from both interior and exterior cameras, learning a unified cross-view representation via a spatio-temporal Transformer to infer driver intention directly from their behavior. However, these methods primarily rely on the Transformer's inherent structure to interpret dependencies. Meanwhile, a non-autoregressive Transformer with hybrid attention has been employed to simultaneously capture the temporal dynamics of a single vehicle and interactions among multiple vehicles \cite{jiang2024hybrid}. DriveTransformer \cite{jiadrivetransformer} further establishes a unified end-to-end framework to handle perception, prediction, and planning tasks in parallel. While this approach treats intention prediction as an integrated component of a scalable system, it may dilute the model's focus on specific driver-intention features.

Besides, some early works also integrated causal inference methods into intention prediction. For instance, one study built a causal model to capture temporal relationships within invariant representations from driving data, aiming for domain generalization \cite{hu2022causal}. Another adopted a driver-centric approach, framing risk object identification as a causal inference problem and introducing a two-stage causal framework \cite{li2020make}. However, these methods typically cover a relatively limited scope of scenarios and conditions.

Building on prior work, we propose a dual‑stream Transformer architecture that leverages a learned intention embedding to explicitly capture both local and global causal dependencies between in‑cabin and external modalities. Through systematic fusion of these multi‑dimensional feature representations, the model achieves highly accurate intention prediction in complex driving environments.

\begin{figure*}[!t]
    \centering
    \includegraphics[width=1.0\linewidth,keepaspectratio]{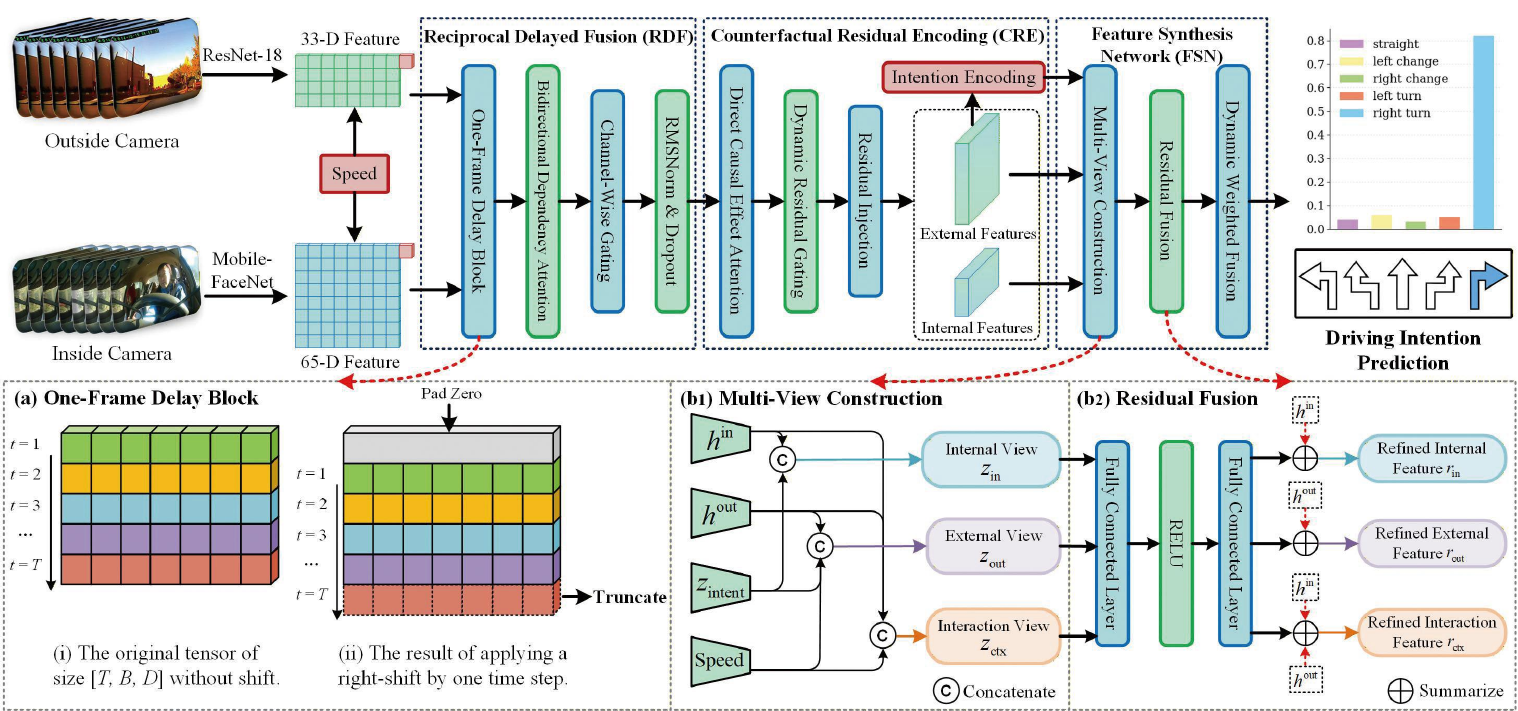}
    \caption{Overview of the \textbf{CaTFormer} pipeline. After data preprocessing, exterior optical flow is encoded by ResNet-18 and interior images by MobileFaceNet to produce dual-stream feature sequences. These are then fed into three core modules: \textbf{(1) Reciprocal Delayed Fusion (RDF)} for temporal feature integration; \textbf{(2) Counterfactual Residual Encoding (CRE)} for causal enhancement and intention embedding; and \textbf{(3) Feature Synthesis Network (FSN)} for dynamic fusion of complementary interior, exterior and interaction views to yield the final driving intention prediction.}
    \label{fig:pipeline}
\end{figure*}

\section{Method}
As illustrated in Fig. \ref{fig:pipeline}, \textbf{CaTFormer} processes a bi‐stream image sequence $\mathcal{I} = \bigl\{(I^{\mathrm{out}}_{b,t}, I^{\mathrm{in}}_{b,t})\bigr\}_{b=1,t=1}^{B,T}$ of $B$ synchronized frame pairs over $T$ time steps, where $b$ and $t$ index the sample and temporal frame, respectively. Feature extractors encode each stream into features $F^{\mathrm{out}}, F^{\mathrm{in}}\in\mathbb{R}^{B\times T\times D}$, where $D$ denotes the dimensionality of each encoded feature vector. The features pass through a Reciprocal Delayed Fusion (RDF) module for temporal causality modeling via our proposed bi-stream attention, followed by a Counterfactual Residual Encoding (CRE) module to inject learnable causal representations. Finally, visual features ($z_{\mathrm{in}}, z_{\mathrm{out}}$ and interaction features $z_{\mathrm{ctx}}$) are adaptively fused by a Feature Synthesis Network (FSN) into a joint prediction $\ell_{\mathrm{joint}}$.

\subsection{Reciprocal Delayed Fusion (RDF)}
On multi‐lane highways, cameras concurrently record the exterior traffic scene and the driver’s interior state. Motivated by bidirectional dependencies between environmental context and driving behavior, our dual‐stream architecture explicitly models their interactions by jointly processing both feature streams.

To model inter-frame temporal precedence, we introduce a temporal delay mechanism in the Key and Value sequences. Specifically, at time step $t$, the attention mechanism accesses only information from the preceding frame $t-1$. Concretely, we define the delayed feature
\begin{equation}
\hat{F}_{b,t} = F_{b,t-1}\,\mathbf{1}_{\{t>1\}},
\quad
\mathbf{1}_{\{t>1\}} =
\begin{cases}
1, & t>1,\\
0, & t=1.
\end{cases}
\end{equation}
applied separately to $F^{\mathrm{out}}$ and $F^{\mathrm{in}}$.
\subsubsection{Bidirectional Dependency Attention (BDA).}
\begin{figure}[!t]
    \centering
    \includegraphics[width=1\linewidth]{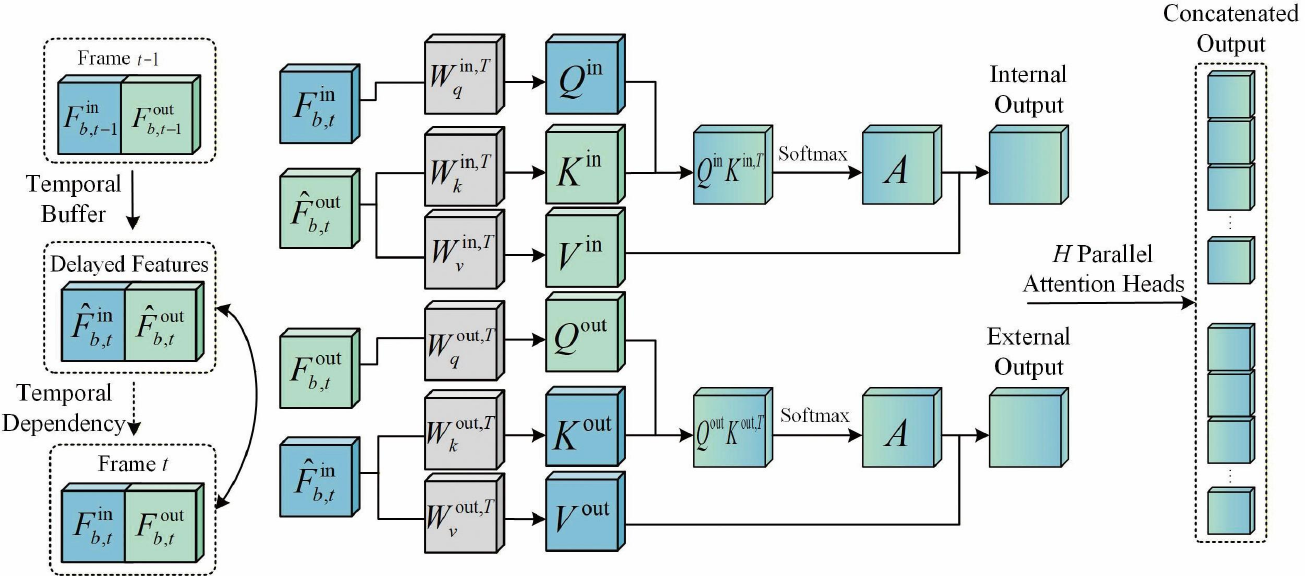}
    \caption{Illustration of Bidirectional Dependency Attention (BDA), where buffered interior and exterior features cross-attend to enhance current-frame representations.}
    \label{fig:cross_attention}
\end{figure}

Under a strict single-frame delay constraint, BDA enriches each frame’s representation by fusing interior and exterior contexts from the immediately preceding timestep. The current interior and exterior features attend bidirectionally to their one-frame delayed counterparts, capturing both temporal coherence and cross-stream coupling. To model diverse associations efficiently, we project into $H$ parallel attention heads (in our experiments, $H=8$) and aggregate their outputs through concatenation and a final linear mapping:
\begin{equation}
\mathrm{BDA}(Q,K,V)
=\Bigl[\mathrm{softmax}\bigl(\tfrac{Q_iK_i^\top}{\sqrt{d_k}}\bigr)V_i\Bigr]_{i=1}^H W^O,
\end{equation}
where $d_k$ is the key dimension, $[\cdot]_{i=1}^H$ denotes concatenation across heads, and $W^O$ restores the original feature size. Fig. \ref{fig:cross_attention} shows the bidirectional query-key-value fusion, highlighting how interior and exterior streams are jointly updated.

\subsubsection{Channel-wise gating.} 
Although BDA generates a fused representation $H_{b,t}$ at each spatial-temporal point, some channels may still carry noise or irrelevant signals. Thus, we apply two channel-wise gating layers to adaptively enhance informative features and suppress spurious ones:
\begin{equation}
\begin{aligned}
g_{b,t}
&= \sigma\bigl(W_{2}\,(\mathrm{ReLU}(W_{1}H_{b,t}+b_{1}))+b_{2}\bigr),
\\
\widetilde H_{b,t}
&= g_{b,t}\,\odot\,H_{b,t},
\end{aligned}
\end{equation}
where $\sigma(\cdot)$ denotes the sigmoid activation function and $\odot$ denotes the Hadamard product between vectors.
\subsubsection{Normalization and regularization.} 
For numerical stability and to guard against overfitting, the following operations are applied to the gated outputs $\widetilde{H}_{b,t}$:
\begin{equation}
    R(x) 
= x \;\odot\; \frac{s}{\sqrt{\frac{1}{D}\sum_{d=1}^D x_d^2 + \epsilon}},
\end{equation}
\begin{equation}
    {X}_{b,t} 
= \mathrm{Dropout}\bigl(R(\widetilde{H}_{b,t})\bigr),
\end{equation}
where $R(\cdot)$ denotes the Root-mean-square normalization function, $s$ is a learnable scaling parameter, $\epsilon$ is a small constant to ensure numerical stability, and $D$ is the dimensionality of the feature vector $x$, yielding channel-calibrated representations across both interior and exterior feature streams.

\subsection{Counterfactual Residual Encoding (CRE)}
Conventional intention prediction architectures aggregate heterogeneous interior and exterior cues under an implicit correlation assumption and thus often mistake coincidental patterns for proper decision drivers. To overcome this limitation, our CRE module contrasts observed and counterfactual cross‐stream attentions to disentangle direct causal contributions. We
 then selectively amplify only those residuals that genuinely influence driving intention, resulting in improved robustness and generalization in safety‐critical scenarios. Specifically, CRE takes the bidirectionally fused interior and exterior features ${X}_{b,T}^{\mathrm{in}}, {X}_{b,T}^{\mathrm{out}}\in\mathbb{R}^{T\times B\times D}$ as inputs to perform this causal reasoning process.
\subsubsection{Direct causal effect.}
At each time step, we compute two attention distributions. We first calculate observed dependency attention $A^{\mathrm{obs}}_{\mathrm{in},t}$ using actual exterior features, and then generate counterfactual dependency attention $A^{\mathrm{cf}}_{\mathrm{in},t}$ by replacing all exterior features with their temporal mean $\bar X^{\mathrm{out}}=\frac{1}{TB}\sum_{u=1}^{T}\sum_{b=1}^{B}X^{\mathrm{out}}_{u,b}$, which serves as a neutral baseline that removes environmental variations. The difference $\Delta^{\mathrm{in}}_{t}$ between these two distributions quantifies the direct causal attention of exterior context on interior representations.
Specifically, for $t=1,\dots,T$, we obtain
\begin{equation}
\begin{aligned}
  A^{\mathrm{obs}}_{\mathrm{in},t}
    &= \mathrm{BDA}\bigl(X^{\mathrm{in}}_{t},\,X^{\mathrm{out}}_{ t-1},\,X^{\mathrm{out}}_{t-1}\bigr),\\
  A^{\mathrm{cf}}_{\mathrm{in},t}
    &= \mathcal{A}\bigl(X^{\mathrm{in}}_{t},\,\bar X^{\mathrm{out}},\,\bar X^{\mathrm{out}}\bigr),\\
  \Delta^{\mathrm{in}}_{t}
    &= A^{\mathrm{obs}}_{\mathrm{in},t} - A^{\mathrm{cf}}_{\mathrm{in},t},
\end{aligned}
\end{equation}
where $\mathcal{A}(Q,K,V)$ denotes multi‐head scaled dot‐product attention. Similarly, $\Delta^{\mathrm{out}}_{t}$ is defined by interchanging the two streams.
We further orthogonalize each causal residual against the global baseline vector $\bar X$ to ensure that the identified causal patterns reflect true intention-relevant dependencies rather than dataset-specific biases. Formally,
\begin{equation}
\Delta_{t}^{\perp}
= \Delta_{t}
- \frac{\Delta_{t}^\top\,\bar {X}}{\|\bar {X}\|^2 + \varepsilon}\,\cdot \bar {X},
\end{equation}
where $\varepsilon$ ensures numerical stability.
The orthogonal projection yields $\Delta^{\perp}_{t}$ by removing baseline-aligned components. We retain $\Delta^{\perp}_{T}$ as the decision-relevant signal, guiding downstream fusion toward temporally salient causal cues.

\subsubsection{Dynamic residual gating.}
The causal relevance of residuals differs across driving scenarios, as critical maneuvers merit amplification, while routine patterns warrant attenuation. We use a learnable gating mechanism to adjust residual contributions according to their predictive value for intention inference.
Specifically, we derive gating coefficients for the orthogonally filtered residuals $\Delta^{\perp,\mathrm{in}}_{T}$ and $\Delta^{\perp,\mathrm{out}}_{T}$ using a linear layer followed by sigmoid activation, and then integrate these gated residuals with the original features:
\begin{equation}
  h^{\mathrm{in}}  = X^{\mathrm{in}}_{T} + g^{\mathrm{in}}_{T} \cdot \Delta^{\perp,\mathrm{in}}_{T},\quad
  h^{\mathrm{out}} = X^{\mathrm{out}}_{T} + g^{\mathrm{out}}_{T} \cdot \Delta^{\perp,\mathrm{out}}_{T},
\end{equation}
where $g^{\mathrm{in}}_{T}$ and $g^{\mathrm{out}}_{T}$ are learned gating coefficients that selectively modulate causal signal contributions to enhance prediction robustness.

\subsubsection{Adaptive intention encoding.} 
Beyond frame-level causal cues, holistic intention understanding requires global semantic reasoning. We extract a coarse intention distribution from the exterior summary $h^{\mathrm{out}}$ through softmax classification over $M$ predefined intention categories ($M$ denotes the total number of classes):
\begin{equation}
\boldsymbol{\xi} = \text{softmax}(W_{\mathrm{int}} h^{\mathrm{out}}) \in \mathbb{R}^{M}
\end{equation}
where $W_{\mathrm{int}}\in\mathbb{R}^{M\times D}$ linearly projects the $D$-dimensional exterior summary onto $M$ logits.
This intention distribution is then re-embedded as an intention token $z_{\mathrm{intent}}\in \mathbb{R}^{D}$ that encodes the driving intention in a continuous representation. The intention token serves as a global semantic anchor, providing top-down guidance across processing streams for consistent interpretation of ambiguous scenarios.

\begin{table*}[t]
  \centering
  {\small
  \setlength\heavyrulewidth{1.5pt}
  \begin{tabular}{cccccccc}
    \toprule
    \textbf{Method}                           & \textbf{Camera} & \textbf{GPS}   & \textbf{Map}   & \textbf{Speed} & \textbf{Pr}   & \textbf{Re}   & \textbf{F1-score} \\
    \midrule
    IOHMM \cite{jain2015carknowsdoanticipating}               
                                     & \checkmark & \checkmark & \checkmark & \checkmark & 74.2 & 71.2 & 72.7  \\
    SDAE \cite{Rekabdar2018}         
                                     & \checkmark &            &            & \checkmark & 71.9 & 74.8 & 73.3  \\
    AIO-HMM \cite{jain2015carknowsdoanticipating}            
                                     & \checkmark & \checkmark & \checkmark & \checkmark & 77.4 & 71.2 & 74.2  \\
    Deep CNN \cite{Rekabdar2018}     
                                     & \checkmark &            &            & \checkmark & 78.0 & 77.5 & 77.7  \\
    FRNN-UL \cite{jain2016RNN}       
                                     & \checkmark &            & \checkmark & \checkmark & 82.2 & 75.9 & 78.9  \\
    FRNN-EL \cite{jain2016RNN}       
                                     & \checkmark &            & \checkmark & \checkmark & 84.5 & 77.1 & 80.6  \\
    FRNN-EL w/ 3D head pose \cite{jain2016RNN} 
                                     & \checkmark &            & \checkmark & \checkmark & 90.5 & 87.4 & 88.9  \\
    LSTM-GRU \cite{Michele2019}      
                                     & \checkmark &            &            & \checkmark & 92.3 & 90.8 & 91.3  \\
    DCNN \cite{Rekabdar2018}         
                                     & \checkmark &            &            & \checkmark & 91.8 & 92.5 & 92.1  \\
    CF-LSTM \cite{Zhou2021}          
                                     & \checkmark &            &            & \checkmark & 92.0 & 92.3 & 92.1  \\
    Predictive-Bi-LSTM-CRF \cite{Zhou2021} 
                                     & \checkmark &            &            &  \checkmark & 92.4 & 94.7 & 93.6  \\
    Central \cite{FedPRM2024} 
                                     & \checkmark & \checkmark &            & \checkmark & 94.4 & 94.3 & 94.2  \\
    FedPRM \cite{FedPRM2024} 
                                     & \checkmark & \checkmark &            & \checkmark & \textbf{99.0} & 92.0 & 95.2  \\                

    \midrule
    Gebert \cite{Patrick2019}        
                                     & \checkmark &            &            &            &    -  &    -  & 81.7  \\
    Rong \cite{Yao2020}              
                                     & \checkmark &            &            &            &    -  &    -  & 84.3  \\
    CEMFormer \cite{ma2023cemformer}             
                                     & \checkmark &            &            &            & -    &   -   & 87.1  \\
    TIFN \cite{TIFN2023}             
                                     & \checkmark &            &            &            & 89.3 & 86.4 & 87.9  \\
    IDIPN \cite{IDIPN2025}           
                                     & \checkmark &            &            &            & 94.2 & 94.9 & 94.5  \\
    \midrule
    \multirow{2}{*}{\textbf{CaTFormer (Ours)}} 
                                     & \checkmark &            &            &            &  96.7    &  \textbf{98.5}    &  97.6     \\
                                     & \checkmark &            &            & \checkmark & 98.7 & \textbf{98.5} & \textbf{98.6}  \\
    \bottomrule
  \end{tabular}}
  \caption{Comparison of state-of-the-art methods on the Brain4Cars dataset using camera and additional sensor modalities (GPS, Map, Speed). The best results are highlighted in bold.}
  \label{tab:result}
\end{table*}

\subsection{Feature Synthesis Network (FSN)}
The CRE module provides a set of disentangled feature vectors corresponding to interior and exterior cues and a preliminary intention token. We further introduce the Feature Synthesis Network (FSN), which performs adaptive fusion of these features to construct a superior synthesized representation for predicting driving intention. By selectively emphasizing the most relevant information, the FSN module enhances the robustness of driving intention prediction.
Each visual branch undergoes a residual nonlinear transformation via a dual-stage feedforward network with intermediate activation, which, combined with the speed feature $s$, yields the fused representations for the interior, exterior, and interaction streams:
\begin{equation}
\begin{aligned}
r_{\mathrm{in}}  &= f_{\mathrm{in}}\bigl([h^{\mathrm{in}},\,z_{\mathrm{intent}}]\bigr) + h^{\mathrm{in}},\\
r_{\mathrm{out}} &= f_{\mathrm{out}}\bigl([h^{\mathrm{out}},\,z_{\mathrm{intent}},\,s]\bigr) + h^{\mathrm{out}},\\
r_{\mathrm{ctx}} &= f_{\mathrm{ctx}}\bigl([h^{\mathrm{in}},\,h^{\mathrm{out}},\,z_{\mathrm{intent}},\,s]\bigr)
           + h^{\mathrm{in}} + h^{\mathrm{out}}
\end{aligned}
\end{equation}
where each $f_\cdot$ denotes a dual-stage feedforward mapping comprising two fully connected layers separated by a ReLU activation (FC–ReLU–FC).
Let $\mathcal{C}=\{\mathrm{in},\mathrm{out},\mathrm{ctx}\}$. Each refined feature $r_i$ ($i\in\mathcal{C}$) is mapped to class logits $\ell_i$ and a corresponding confidence weight $w_i$, which adaptively controls each branch's contribution:
\begin{equation}
w_i = \frac{\exp(u_i^\top r_i)}{\sum_{j\in\mathcal{C}}\exp(u_j^\top r_j)} ,\quad \ell_{\mathrm{joint}} = \sum_{i\in\mathcal{C}} w_i\, (W_i\,r_i).
\end{equation}

\subsection{Model Training}
To address class imbalance and enhance sensitivity to rare intentions while promoting early prediction, we design a unified loss function that combines the average cross-entropy (CE) across complementary streams with an intention-prediction term:
\begin{equation}
  \mathcal{L}
  = \underbrace{\frac{1}{4}
    \sum_{i\in\mathcal{H}}
    \mathrm{CE}\bigl(\ell_i,\,y\bigr)
  }_{\text{main loss}}
  \;+\;
  \underbrace{\alpha\,
    \mathrm{CE}\bigl(\ell_{\mathrm{intent}},\,y\bigr)
  }_{\text{intention loss}}
\end{equation}
where $\mathcal{H} = \{\mathrm{in}, \mathrm{out}, \mathrm{ctx}, \mathrm{joint}\}$ denotes the four stream-level heads, and $\alpha$ controls the weight of the intention supervision term.
This unified objective integrates class-imbalance mitigation, multi-view fusion, and intention supervision within a cohesive framework.  

\section{Experiments}
\subsection{Data Preparation}
\textbf{\textit{Brain4Cars:}} The Brain4Cars dataset \cite{jain2016brain4carscarknowssensoryfusion} comprises exterior (480 $\times$ 720) and interior (1088 $\times$ 1920) videos of up to 5-second segments, refined to 594 valid events after excluding incomplete or unsynchronized samples. Each video is uniformly sampled to 150 frames, extracting the 5-second segment preceding the maneuver. Interior frames are cropped to 900 $\times$ 800, resized to 112 $\times$ 112, and encoded by a MobileFaceNet yielding 64-D features. Exterior frames undergo Farneback optical flow computation, are resized to 144 $\times$ 96, and then processed by ResNet-18 to produce 32-D features. Appending a smoothed speed signal yields 65-D (interior) and 33-D (exterior) vectors. These vectors are linearly projected, positionally encoded, and passed through a Transformer encoder to obtain temporal representations for CaTFormer. The dataset spans highway and urban settings with five maneuver classes: straight, left turn, right turn, left lane change, and right lane change.

\begin{table}[!t]
  \centering
  {\small
    \setlength\heavyrulewidth{1.5pt}
    \begin{tabular}{%
      >{\centering\arraybackslash}m{2.2cm}
      >{\centering\arraybackslash}m{0.7cm}
      >{\centering\arraybackslash}m{0.7cm}
      >{\centering\arraybackslash}m{1.0cm}
      >{\centering\arraybackslash}m{1.5cm}
    }
      \toprule
      \textbf{Method} & \textbf{In} & \textbf{Out} & \textbf{F1 (\%)} & \textbf{Param.\,(M)} \\
      \midrule
      \multirow{3}{*}{\parbox{2.2cm}{\centering Gebert \cite{Patrick2019}}}
        & $\checkmark$ &            & 81.7 & 85.3+162  \\
        &              & $\checkmark$ & 43.4 & 85.3+162  \\
        & $\checkmark$ & $\checkmark$ & 73.2 & 170.5+162 \\
      \midrule
      \multirow{3}{*}{\parbox{2.2cm}{\centering Rong \cite{Yao2020}}}
        & $\checkmark$ &            & 75.5 & 46.2+162  \\
        &              & $\checkmark$ & 66.4 & 5.4+162   \\
        & $\checkmark$ & $\checkmark$ & 84.3 & 57.9+162  \\
      \midrule
      TIFN \cite{TIFN2023}         & $\checkmark$ & $\checkmark$ & 87.9 & 12.3+5.3  \\
      \midrule
      IDIPN \cite{IDIPN2025}       & $\checkmark$ & $\checkmark$ & 94.5 & \textbf{11.75+5.3} \\
      \midrule
      \textbf{CaTFormer (Ours)}   & $\checkmark$ & $\checkmark$ & \textbf{98.6} & 14.53+5.3 \\
      \bottomrule
    \end{tabular}
    \caption{Comparison of our CaTFormer against other end-to-end methods on the Brain4Cars dataset, using interior and exterior streams, with F1-score (\%) and parameters (M).}
    \label{tab:comparison}
  }
\end{table}

\subsection{Implementation Details}
Our proposed CaTFormer was implemented by PyTorch, and experiments were performed on a server with six NVIDIA RTX 2080 Ti GPUs. The model was trained end‐to‐end on Brain4Cars using the Adam optimizer (initial learning rate $1\times10^{-3}$) for 160 epochs with a batch size of 16. During training, each input comprised a chunk of frames randomly sampled from the 5-second pre-maneuver segment. When testing, chunks were obtained via uniform sampling. The weight $\alpha$ in the unified loss was empirically set to 0.1. Model performance was evaluated using 5-fold cross-validation.

\subsection{Evaluation Protocols}
In driving intention prediction, straight driving is considered background, and only turns and lane changes are treated as target events. To evaluate our CaTFormer model, we define the following prediction-based metrics: true positives (TP: correctly predicted maneuvers), false positives (FP: maneuvers misclassified as another maneuver), false optimistic predictions (FPP: predicting a maneuver when none occurred), and missing predictions (MP: failing to detect an actual maneuver). Given the set of all behaviors $\mathcal G$ and target maneuvers $\mathcal G'=\mathcal G\setminus\{\mathrm{straight}\}$, Precision (Pr), Recall (Re), and F1-score are computed as follows:
\begin{equation}
\begin{aligned}
    \mathrm{Pr}&=\frac{1}{|\mathcal G'|}\sum_{m\in\mathcal G}\frac{TP_m}{TP_m+FP_m+FPP_m},\\
\mathrm{Re}&=\frac{1}{|\mathcal G'|}\sum_{m\in\mathcal G}\frac{TP_m}{TP_m+MP_m},\ F_{1}=\frac{2*\mathrm{Pr}*\mathrm{Re}}{\mathrm{Pr}+\mathrm{Re}}.
\end{aligned} 
\end{equation}

\begin{table}[!t]
  \centering
  \setlength{\tabcolsep}{2pt}
  {\small
    \begin{tabular}{
      >{\centering\arraybackslash}m{2.5cm}
      *{5}{>{\centering\arraybackslash}m{0.91cm}}
    }
      \toprule[1.5pt]
      \multirow{2}{*}{\textbf{Method}}
        & \multicolumn{5}{c}{\textbf{F1-score (\%)}} \\
      \cmidrule[0.2pt](lr){2-6}
        & [-5,0] & [-5,-1] & [-5,-2] & [-5,-3] & [-5,-4] \\
      \midrule
      Rong \cite{Yao2020} (in)        & 75.7 & 73.1 & 68.6 & 58.5 & 48.2 \\
      Rong \cite{Yao2020} (out)       & 66.4 & 62.4 & 47.0 & 38.8 & 38.9 \\
      Rong \cite{Yao2020} (both)      & 84.3 & 78.9 & 70.6 & 60.3 & 53.4 \\
      TIFN \cite{TIFN2023}            & 87.9 & 80.9 & 71.0 & 55.0 & 44.6 \\
      IDIPN \cite{IDIPN2025}          & 94.5 & 84.1 & 74.2 & 62.0 & 55.4 \\
      \textbf{CaTFormer (Ours)}      & \textbf{98.6} & \textbf{97.4} & \textbf{90.1} & \textbf{78.4} & \textbf{63.7} \\
      \bottomrule[1.5pt]
    \end{tabular}
  }
  \caption{F1-scores on the Brain4Cars dataset for evaluation on video segments truncated 1–4 s before action onset.}
  \label{table:evaluate_incomplete}
\end{table}

\subsection{Comparison with State-of-the-art Methods}
\begin{figure}[!t]
    \centering
    \includegraphics[width=1\linewidth]{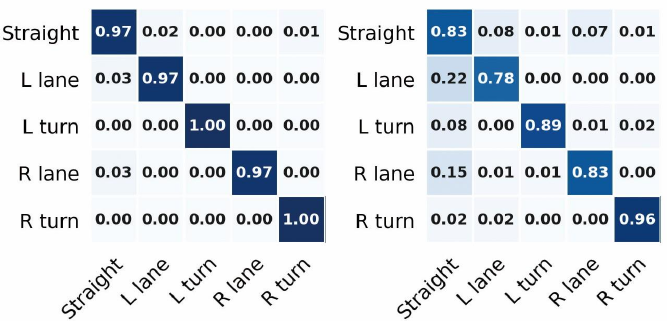}
    \caption{The confusion matrix tested on Brain4cars dataset. Left is ours, right is the result of TIFN \cite{TIFN2023}. The color deepens as the value increases.}
    \label{fig:cmp_TIFN}
\end{figure}

Table \ref{tab:result} provides a systematic comparison of both single- and multi-modal methods on the Brain4Cars dataset. Notably, our camera-only CaTFormer variant achieves an F1-score of 97.6\% (precision 96.7\%, recall 98.5\%), markedly surpassing all previous single-modality methods such as DCNN (92.1\%) and CF-LSTM (92.1\%). When enriched with speed information, CaTFormer attains a new state-of-the-art F1-score of 98.6\% (precision 98.7\%, recall 98.5\%), outperforming the best prior multi-modal model, FedPRM (95.2\% F1), by 3.4\%. These results demonstrate that CaTFormer not only establishes a new performance standard but does so with fewer sensor inputs, highlighting its efficiency and robustness in driving intention prediction.
Fig. \ref{fig:cmp_TIFN} presents the confusion matrices of our CaTFormer and TIFN \cite{TIFN2023}. CaTFormer yields a sharper diagonal and substantially fewer off-diagonal entries, demonstrating its superior discrimination of similar maneuvers and reduced false predictions.
Detailed comparative results between our method and other end-to-end approaches on full-video inputs appear in Table \ref{tab:comparison}. As the optical-flow algorithm lies outside the core prediction pipeline, its parameters are listed separately. Our model achieves superior recognition performance with only a marginal increase in model size, demonstrating its compact efficiency.

In addition to evaluating F1-scores on complete videos (–5 s to 0 s, where 0 s marks driver action), we assessed early-warning capability by truncating observation windows at –1 s, –2 s, –3 s, and –4 s. As shown in Table \ref{table:evaluate_incomplete}, prediction accuracy declines nearly linearly with shorter observations, highlighting increased uncertainty at longer forecast horizons. This result reflects the intrinsic trade-off between early intervention and predictive accuracy. Our CaTFormer consistently achieves superior performance across all truncated settings, demonstrating its robustness in driving intention prediction.

\begin{table}[!t]
  \centering
  {\small
    \begin{tabular}{%
      >{\centering\arraybackslash}m{0.18\columnwidth}
      *{5}{>{\centering\arraybackslash}m{0.10\columnwidth}}
    }
      \toprule[1.5pt]
      \multirow{2}{*}{\textbf{Model}} & \multicolumn{5}{c}{\textbf{F1‐score (\%)}} \\ 
      \cmidrule(lr){2-6}
                             & [-5,0] & [-5,-1] & [-5,-2] & [-5,-3] & [-5,-4] \\
      \midrule
      Base                   & 95.8 & 94.2 & 85.4 & 73.7 & 63.2 \\
      Base+R                 & 97.1 & 95.6 & 87.1 & 75.2 & 61.9 \\
      Base+C                 & 97.0 & 95.4 & 86.9 & 74.9 & 61.1 \\
      Base+F                 & 96.6 & 94.9 & 86.3 & 73.9 & 62.6 \\
      Base+R+C               & 97.4 & 95.7 & 87.4 & 75.9 & 60.3 \\
      Base+R+F               & 98.0 & 96.7 & 88.5 & 76.8 & \textbf{65.6} \\
      Base+C+F               & 97.8 & 96.4 & 88.0 & 76.2 & 62.4 \\
      CaTFormer (R+C+F)      & \textbf{98.6} & \textbf{97.4} & \textbf{90.1} & \textbf{78.4} & 63.7 \\
      \bottomrule[1.5pt]
    \end{tabular}
   }
   \caption{F1‐scores on the Brain4Cars dataset for the dual-stream Transformer baseline (\textbf{Base}) and its variants augmented with RDF (\textbf{R}), CRE (\textbf{C}), and FSN (\textbf{F}).}
   \label{tab:module_ablation}
\end{table}

\begin{figure*}
    \centering
    \includegraphics[width=1\linewidth]{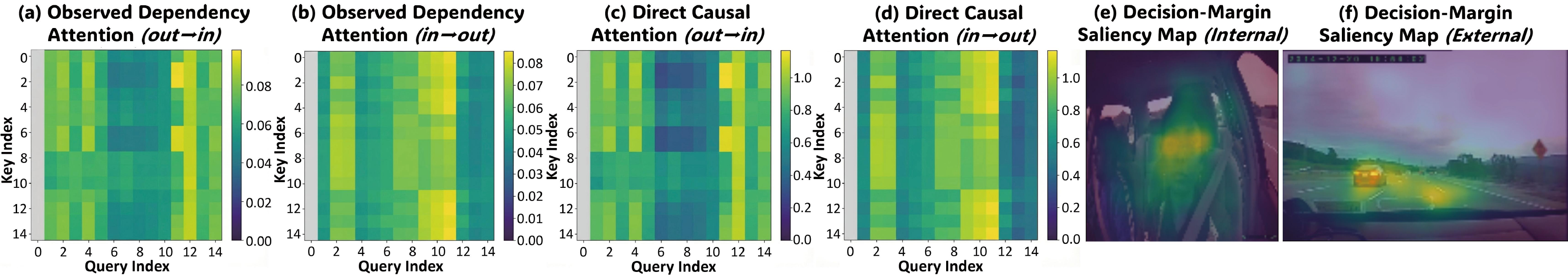}
    \caption{Temporal Attention and Decision‑Margin Saliency Visualizations. (a) and (b) display the \textit{Observed Dependency Attention}, where each pixel $w_{ij}$ (row $i$, column $j$) represents the attention weight from the $j$-th query to the $i$-th key for out$\to$in and in$\to$out directions, respectively. (c) and (d) demonstrate the \textit{Direct Causal Attention}, highlighting frames with significant causal influence. The first column is gray masked to mark shift padding. (e) and (f) overlay the Decision‑Margin Saliency Map on the final interior and exterior frames, each pixel’s intensity defined by $\sum\limits_{t\in\mathcal{T}}\alpha_t\Bigl\lvert\partial\bigl(z_{c^*}- \tfrac{1}{\mathcal{G}-1}\sum\limits_{c\neq c^*} z_c\bigr)/\partial x_t\Bigr\rvert$, where $\alpha_t$ denotes the causal–attention weight for frame $t$, $x_t$ denotes the input feature vector at that pixel, $z_c$ is the final logit for class $c$ and $c^*$ is the predicted class, highlighting regions most responsible for the model’s final decision. Yellow indicates stronger attention.}
    \label{fig:attention_visual}
\end{figure*}

\subsection{Result Visualization}
Fig. \ref{fig:attention_visual} demonstrates that the model employs a temporal attention mechanism to realize a full reasoning path from dynamic event understanding to static decision attribution. Temporally, (a) and (b) delineate broad, task‑relevant event windows (e.g., the lane‑change interval), whereas (c) and (d) concentrate on a small set of decisive frames, accentuating discriminative cues and pinpointing instantaneous triggers. This sequence closely mirrors human cognition in which one perceives an event in its entirety before pinpointing its core cause. In the spatial dimension, (e) and (f) anchor the model’s reasoning in semantically relevant regions, including the driver’s facial state inside the cabin and the key road environment outside, entirely consistent with real‑world logic. This process vividly demonstrates how the model integrates critical cues to arrive at a reliable judgment, thereby substantiating the soundness of its decision‑making.

\subsection{Ablation Study}
\subsubsection{Effect of components in CaTFormer.} 
To evaluate the contribution of each component in CaTFormer, we conduct systematic ablation studies, as shown in Table~\ref{tab:module_ablation}. Starting from a dual-stream Transformer baseline (Base) that performs late fusion via feature concatenation, we progressively add RDF, CRE, and FSN to measure their impact on F1-score. We further explored various combinations to explore module interactions and their cumulative effects. Experimental results confirm that each module improves intention prediction, with additional gains from their integration.

\begin{figure}[!t]
    \centering
    \includegraphics[width=1\linewidth]{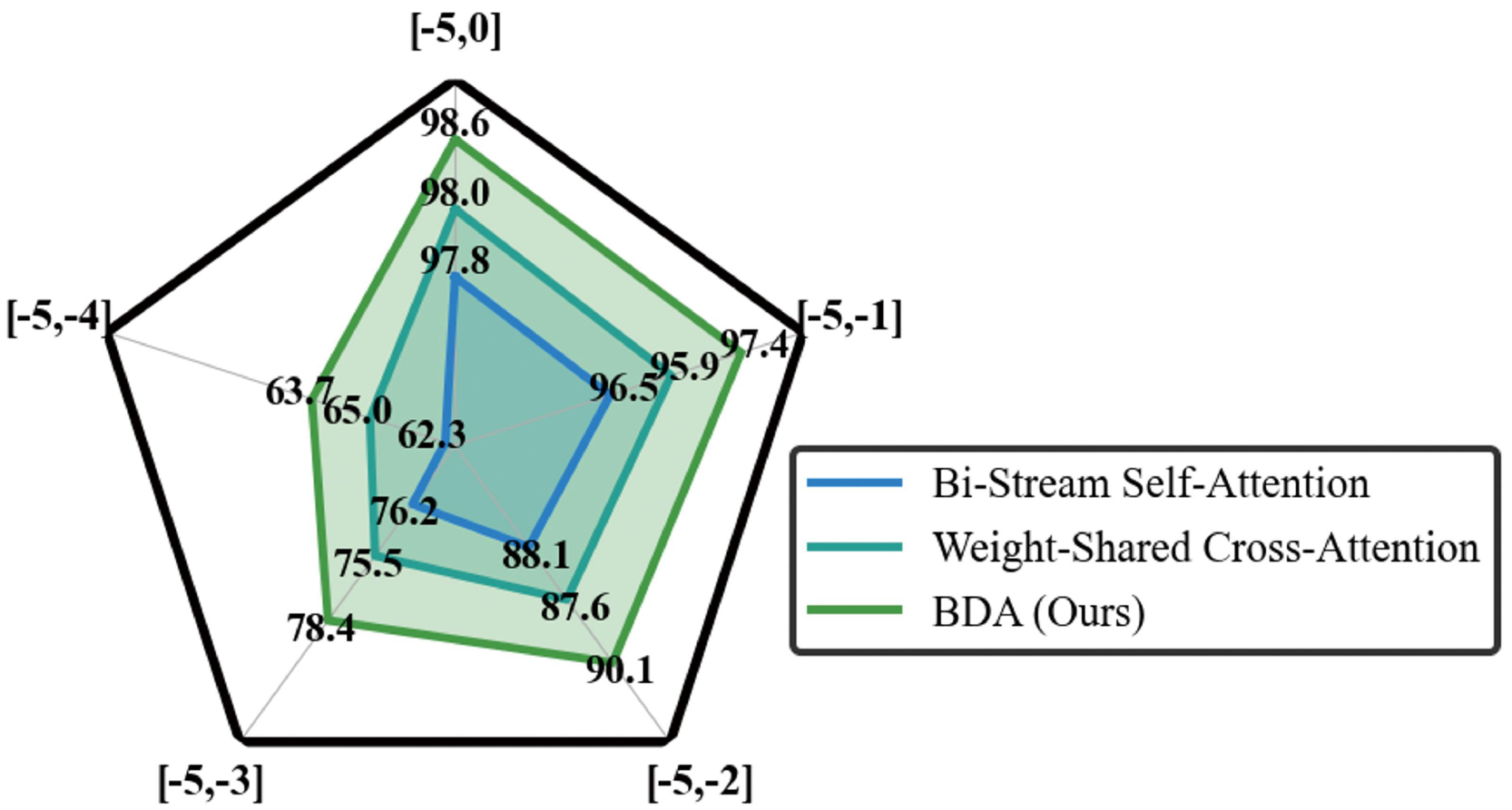}
    \caption{F1-scores (\%) of various attention mechanisms on the Brain4Cars dataset, offset by 15 units for visual clarity.}
    \label{fig:attention_ablation}
\end{figure}

\subsubsection{Effect of attention design in CaTFormer.}
To evaluate the effect of different attention mechanisms on driving intention prediction, we conducted ablation experiments on the Brain4Cars dataset comparing three attention schemes, as summarized in Fig. \ref{fig:attention_ablation}. The results indicate that our proposed Bidirectional Dependency Attention (BDA) more effectively captures spatio-temporal correlations between interior and exterior streams while suppressing noise, thereby demonstrating its superiority in isolating dynamic cues and enhancing overall model robustness.

\begin{figure}[!t]
    \centering
    \includegraphics[width=1\linewidth]{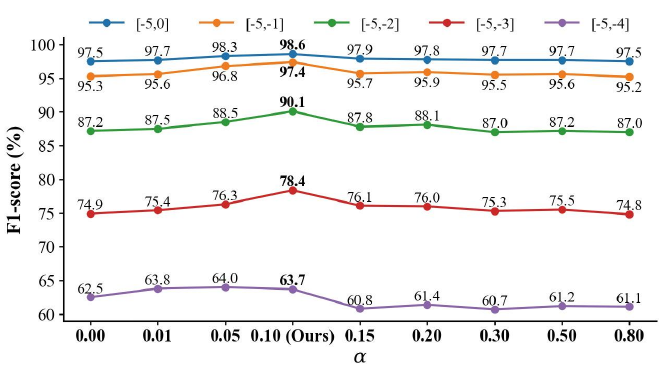}
    \caption{F1‐scores (\%) for different values of $\alpha$ in the loss function on the Brain4Cars dataset.}
    \label{fig:loss_ablation}
\end{figure}

\subsubsection{Effect of intention loss.}
We study the effect of the intention-loss weight $\alpha$ through an ablation analysis. As shown in Fig.~\ref{fig:loss_ablation}, $\alpha=0.10$ achieves the best F1-score, balancing temporal learning with intention supervision and providing informative gradients without destabilizing the main objective. Smaller $\alpha$ under-supervises subtle pre-action cues, whereas larger $\alpha$ induces gradient conflicts that degrade temporal coherence and anticipation accuracy.

\section{Conclusion}
In this paper, we introduced CaTFormer, a unified architecture that explicitly models causal interactions between driver behavior and environmental context for accurate intention prediction. Our approach extracts dual-stream features and merges them through a structured fusion pipeline. Extensive experiments on the Brain4Cars dataset confirm that CaTFormer achieves state-of-the-art accuracy, demonstrating its suitability for real-time driver assistance.

\section*{Acknowledgments}
This work was supported in part by the Beijing Jiaotong University Research Fund under Grant KKA309004533.

\bibliography{aaai2026}

\end{document}